\documentclass[11pt,a4paper]{article}
\usepackage[hyperref]{acl2021}
\usepackage{times}
\usepackage{latexsym}

\usepackage{amsmath,amssymb,graphicx,wasysym}
\usepackage{multirow,booktabs,xcolor,colortbl}
\usepackage{tabularx}
\graphicspath{{figures/}}
\usepackage{mathtools}
\usepackage{microtype}
\usepackage{fancybox}

\aclfinalcopy % Uncomment this line for the final submission
\providecommand\longrightarrowRHD{\relbar\joinrel\relbar\joinrel\mathrel\RHD}
\providecommand\longrightarrowrhd{\relbar\joinrel\relbar\joinrel\mathrel\rhd}

\makeatletter
\providecommand*\xrightarrowRHD[2][]{\ext@arrow 0055{\arrowfill@\relbar\relbar\longrightarrowRHD}{#1}{#2}}
\providecommand*\xrightarrowrhd[2][]{\ext@arrow 0055{\arrowfill@\relbar\relbar\longrightarrowrhd}{#1}{#2}}
\makeatother

\def\ie{i.e.\ }

\newcommand\unc{$^\bigstar$}
\newcommand\bs{$^\ddagger$}
\newcommand\ra{$\rightarrow$}
\newcommand{\MC}[3]{\multicolumn{#1}{#2}{#3}}

\newcommand{\B}{\textbf}
\newcommand{\I}{\textit}
\newcommand{\T}{\texttt}

\newcommand{\model}{\textsc{BertGen}}
\newcommand{\mbert}{\textsc{M-Bert}}
\newcommand{\vlbert}{\textsc{VL-Bert}}

\title{\model{}: Multi-task Generation through BERT}

\author{\\
	\textbf{Faidon Mitzalis\textsuperscript{1},  Ozan Caglayan\textsuperscript{1}, Pranava Madhyastha\textsuperscript{1}, Lucia Specia\textsuperscript{1,2}}\\
	\textsuperscript{1}Department of Computing, Imperial College London, UK \\
	\textsuperscript{2}Department of Computer Science, University of Sheffield, UK\\
	\texttt{phaedonmit@gmail.com, \{o.caglayan,pranava,lspecia\}@ic.ac.uk}\\
}

\date{}

\begin{document}
\maketitle

\begin{abstract}
We present \model{}, a novel generative, decoder-only model which extends BERT by fusing multimodal and multilingual pre-trained models \vlbert{} and \mbert{}, respectively. \model{} is auto-regressively trained for language generation tasks, namely image captioning, machine translation and multimodal machine translation, under a multi-task setting.
With a comprehensive set of evaluations, we show that \model{} outperforms many strong baselines across the tasks explored. We also show \model{}'s ability for zero-shot language generation, where it  exhibits competitive performance to supervised counterparts.
Finally, we conduct ablation studies which demonstrate that \model{} substantially benefits  from multi-tasking and effectively transfers relevant inductive biases from the pre-trained models.
\end{abstract}

%%%%%%%%%%%%%%%%%%%%%%
\section{Introduction}
\label{sec:intro}
Recent work in unsupervised and self-supervised pre-training has revolutionised the field of natural language understanding (NLU), resulting in high performance ceilings across multiple tasks~\cite{devlin-etal-2019-bert,yang2019xlnet,dong-etal-2019-unilm}.
The recent success of language model pre-training with masked language modelling (MLM) such as BERT~\cite{devlin-etal-2019-bert} further paved the way for more complex approaches that combine language pre-training with images~\cite{tan-bansal-2019-lxmert,su2019vl,Lu_2020_CVPR}, video~\cite{sun2019videobert}, and speech~\cite{chuang2019speechbert}.
Most of these approaches follow a task-specific fine-tuning step after the model is pre-trained.

However, there has been little work on exploiting pre-trained MLMs for natural language generation (NLG) tasks. Previous work argues that the MLM objective is ill-suited for generation tasks such as machine translation~\cite{yang2019xlnet,rothe-etal-2020-leveraging}. Recent work in this direction has predominantly investigated the use of pre-trained models to either initialise Transformer-based encoder-decoder models~\cite{imamura-sumita-2019-recycling,clinchant-etal-2019-use,yang2020towards,rothe-etal-2020-leveraging} or to distill knowledge for sequence generation tasks~\cite{chen-etal-2020-distilling}.

In this work, we present \model{}, which extends BERT in a generative setting ($\S$~\ref{sec:arch}). This results in a single generator -- without a separation between the encoder and the decoder -- capable of consuming multiple input modalities and generating in multiple languages. The latter features are achieved by transferring knowledge from state-of-the-art pre-trained models, namely \vlbert{}~\cite{su2019vl} and multilingual BERT (\mbert{})~\cite{devlin-etal-2019-bert}. We train \model{} on various tasks, including image captioning, machine translation and multimodal machine translation, and datasets in four different languages ($\S$~\ref{sec:data_tasks}).

Based on a number of experiments, our findings ($\S$~\ref{sec:results}) show that \model{} (i) is surprisingly versatile as it is capable of describing images and performing translation in unimodal and multimodal settings, across all languages, (ii) generalises well across zero-shot image captioning, multimodal machine translation, and out-of-domain news translation tasks, and finally (iii)
is parameter efficient when compared to state-of-the-art models for each of the tasks combined together.
%%%%%%%%%%%%%%%%%%%%%%

%%%%%%%%%%%%%%%%%%%%%%%%%%%%%%%
\section{Method}
\label{sec:method}
%%%%%%%%%%%%%%%%%%%%%%%
\begin{figure*}[t!]
\centering
\includegraphics[width=.98\textwidth]{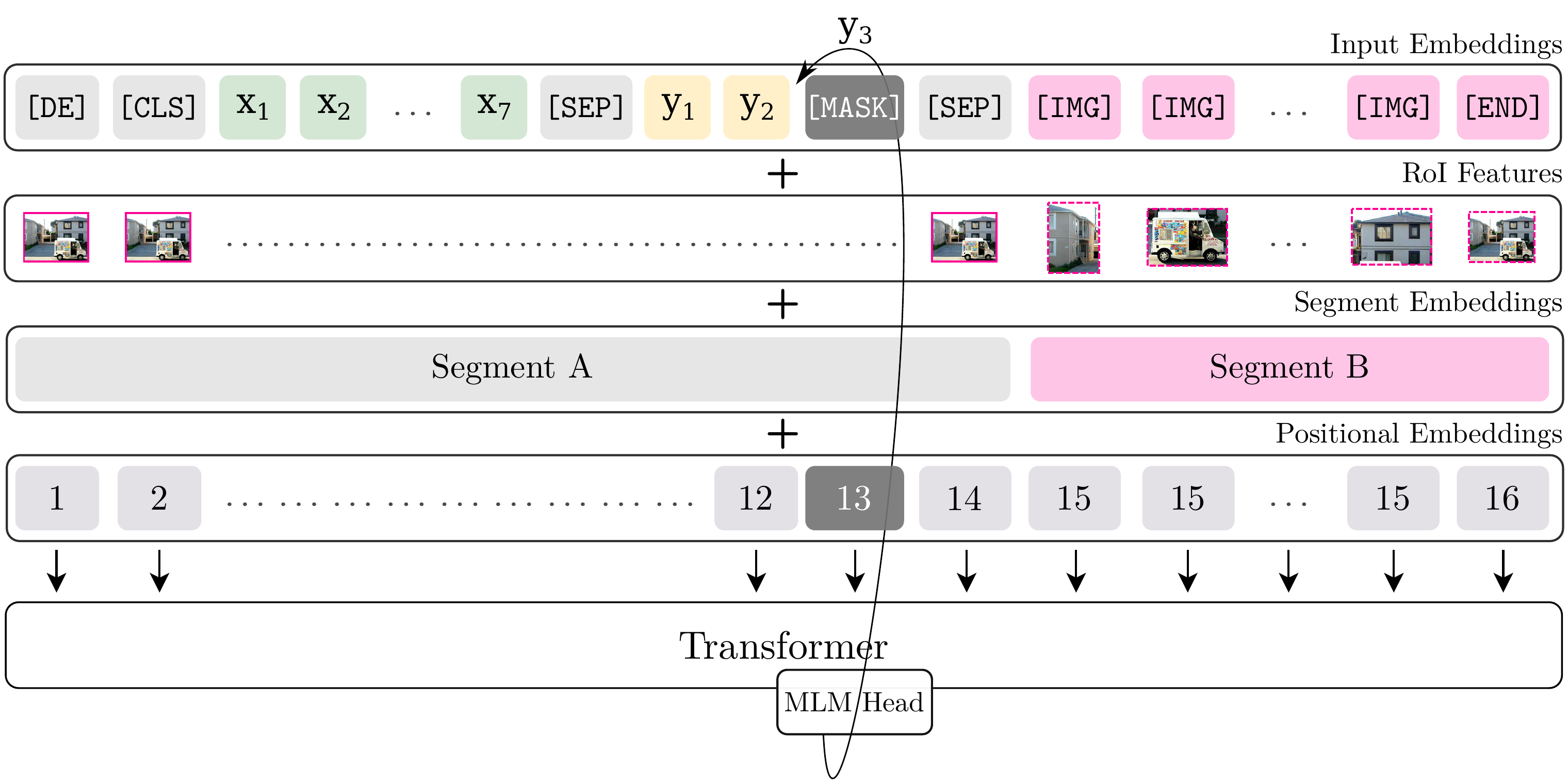}
\caption{A view of the \model{} model during the training of an MMT sample: solid and dashed borders around the images represent full-image features and regional features, respectively. At \B{test time}, the most likely token $y_3 = \text{argmax}(P(y_t | \mathbf{x,v,y_{<t}}))$ is placed back into the sequence and the \T{[MASK]} token is shifted right by one.}
\label{fig:mmbert}
\end{figure*}
%%%%%%%%%%%%%%%%%%%%%%%
In this section, we describe \model{} and the tasks  we explore. We then detail the baselines and SoTA systems that we compare against.

%%%%%%%%%%%%%%%%%%%%%%%%%%%%%%%%%%%%
\subsection{Model}
\label{sec:arch}
%%%%%%%%%%%%%%%%%%%%%%%%%%%%%%%%%%%%
This section details the main aspects of \model{} that distinguish it from the existing work on vision \& language pre-training.

\begin{figure}[t]
 \centering 
    \includegraphics[width=.4\textwidth]{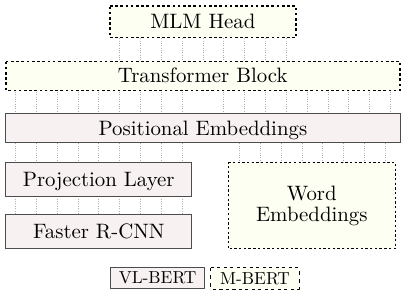}
    \caption{Hybrid initialisation: the \B{solid} and \B{dashed} blocks are transferred from pre-trained \vlbert{} and \mbert{} checkpoints, respectively.}
    \label{fig:hybrid}
\end{figure}

%%%%%%%%%%%%%%%%%%%%%%%%%%%
\paragraph{Initialisation.}
\label{sec:method_init}
%%%%%%%%%%%%%%%%%%%%%%%%%%%
% Done
We take advantage of the previous successes in large-scale pre-training and propose a hybrid initialisation for \model{} (Figure~\ref{fig:hybrid}). This involves using the \vlbert{}~\cite{su2019vl} checkpoint
%\footnote{https://github.com/jackroos/VL-BERT}
and initialising the word embeddings, the Transformer weights and the MLM head with \mbert{}~\cite{devlin-etal-2019-bert}.
We conjecture that this primes \model{} to be aware of the visual modality and of multiple languages. This is simply due to \vlbert{} being pre-trained on English monolingual and image captioning corpora, as well as \mbert{} offering a 119K WordPiece vocabulary, trained on the entire Wikipedia in 104 languages\footnote{We adopt the `BERT-Base, Multilingual Cased' version from the Transformers toolkit~\cite{wolf-etal-2020-transformers}.}.

%%%%%%%%%%%%%%%%%%%%%%%%%%%%%%%%
\paragraph{Input configuration.}
%%%%%%%%%%%%%%%%%%%%%%%%%%%%%%%%
% Done
While \model{} is potentially capable of modeling a variety of generative tasks, we focus on three particular tasks, namely machine translation (MT), multimodal MT (MMT) and image captioning (IC). Therefore, depending on the task, the input configuration of the model may change during both training and testing. To clarify further,
let us first denote a sequence of embeddings representing a source sentence by $\mathbf{x}^{(i)} = [x^{(i)}_1,\cdots,x^{(i)}_m]$, its target translation by $\mathbf{y}^{(i)} = [y^{(i)}_1,\cdots,y^{(i)}_n]$, and a collection of $k$ regional visual features extracted from an associated image by $\mathbf{v}^{(i)}=[v^{(i)}_1,\cdots,v^{(i)}_k]$.
Figure~\ref{fig:mmbert} depicts \model{} when processing a sample from the MMT task. This task's input configuration is a triplet that involves all the three sequences \ie $\{\mathbf{x}^{(i)}, \mathbf{y}^{(i)}, \mathbf{v}^{(i)}\}$. Using this notation, the MT and IC tasks' configurations would correspond to $\{\mathbf{x}^{(i)}, \mathbf{y}^{(i)}\}$ and $\{\mathbf{v}^{(i)}, \mathbf{y}^{(i)}\}$, respectively.

%%%%%%%%%%%%%%%%%%%%%%%%%%%
\paragraph{Visual embeddings.}
%%%%%%%%%%%%%%%%%%%%%%%%%%%
% Done
We follow \vlbert{} and represent images as a collection of $k$ features $\mathbf{v}^{(i)}$ defined for regions of interest (RoI). After pre-extracting the 2048-dimensional RoI features using the \I{bottom-up-top-down} object detector~\cite{anderson2018bottom},
we keep between 10 and 100 (\ie $k \in \left[10,100\right]$) of them depending on the confidence score. The final visual embedding for an RoI is obtained by summing its feature vector and its geometric embedding (\ie the projection of the bounding box coordinates). When encoding the non-visual positions, the same RoI feature vector for the full image is repeated (see  Figure~\ref{fig:mmbert}). We note that we do \B{not fine-tune} the object detector during training.
\begin{figure}[t]
 \centering 
    \includegraphics[width=.45\textwidth]{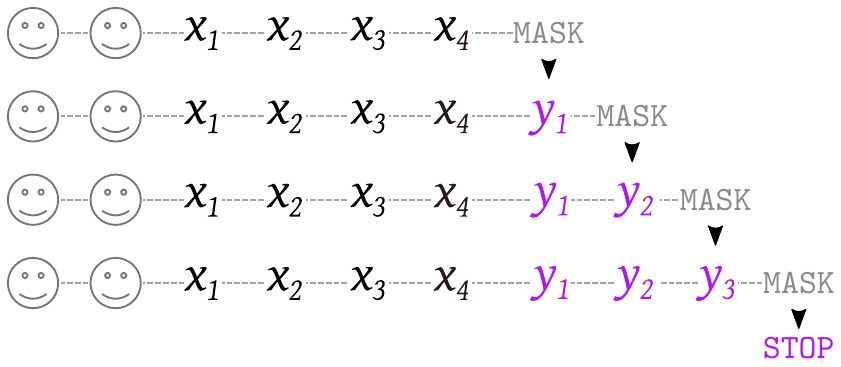}
    \caption{A look at \model{}'s self-attention: the connections denote that self-attentive representations are re-computed in every step. The generation ends when \T{STOP} is predicted. The smileys refer to RoI features.}
    \label{fig:masks}
\end{figure}

%%%%%%%%%%%%%%%%%%%%%%%%%%%%%%%
\paragraph{Sequence unrolling.}
%%%%%%%%%%%%%%%%%%%%%%%%%%%%%%%
An important aspect of \model{} is that it does not explicitly distinguish between the encoder and the decoder blocks usually seen in sequence-to-sequence models. This is accomplished by formalising both encoding and generation using the MLM framework. Formally, let us consider the MMT task and define the maximum log-likelihood objective for a given triplet $\{\mathbf{x}^{(i)}, \mathbf{v}^{(i)}, \mathbf{y}^{(i)}\}$ where the target $\mathbf{y}^{(i)}$ has $n$ tokens:
\begin{align}
    \mathcal{L}^{(i)} = \sum_{t=1}^{n}\log\left(P(y^{(i)}_t \,|\,  \mathbf{x}^{(i)}; \mathbf{v}^{(i)}; \mathbf{y^{(i)}_{<t}})\right)\label{eq:ce}
\end{align}
In a typical sequence-to-sequence model, each log-probability term would be computed by a \I{decoder} within the forward-pass of the \I{same} training example. In contrast, \model{} explicitly unrolls the example $n$ times, forming $n$ new training examples. In other words, each conditional term in Equation~\ref{eq:ce} is observed independently within an epoch of training. Therefore, sequence unrolling has a data \B{augmentation} effect since a training corpus with $D$ examples is approximately augmented by a factor of the average length of the target sequences. Moreover, the unified encoder-decoder formalism halves the number of parameters, making \model{} parameter efficient.

%%%%%%%%%%%%%%%%%%%%%%%%%%%
\paragraph{Self attention.}
\label{sec:self_att}
%%%%%%%%%%%%%%%%%%%%%%%%%%%
Given that a single Transformer~\cite{vaswani2017attention} performs both encoding and decoding, sequence unrolling affects \B{self-attention} as well (Figure~\ref{fig:masks}). First, all positions attend to each other for a given unrolled example \ie the attention is bi-directional. Second, since each unrolled case is an independent example, the self-attentive representations of early positions are naturally \B{re-computed}, in contrast to typical Transformer decoders. Finally, due to how inputs/outputs are represented in a single stream and encoded through shared self-attention, \model{} enforces an inductive bias towards a truly multi-modal and multi-lingual representation space.

%%%%%%%%%%%%%%%%%%%%%%%%%%%%%%%%%%%%%%%
\paragraph{Target language specifiers.}
%%%%%%%%%%%%%%%%%%%%%%%%%%%%%%%%%%%%%%%
Finally, to select the language during generation, input sequences begin with special target language specifiers~\cite{ha-etal-2016-toward,johnson-etal-2017-googles} (Figure~\ref{fig:mmbert}). The specifier is task-agnostic, \ie the same specifier \T{[DE]} is used both when captioning into German and when translating into German.

%%%%%%%%%%%%%%%%%%%%%%%%%%%%%%%%%%%%%%%%
\paragraph{Training \& hyper-parameters.}
%%%%%%%%%%%%%%%%%%%%%%%%%%%%%%%%%%%%%%%%
We extend\footnote{\href{https://github.com/ImperialNLP/BertGen}{https://github.com/ImperialNLP/BertGen}} the base configuration of
\vlbert{} which is a Transformer with 12 self-attention layers and 12 heads. The model and feed-forward dimensions are 768 and 3072, respectively. On a single 32GB V100 GPU, one epoch ($\S$~\ref{sec:results}) takes approximately two days to complete as we could only fit one example per task (\ie batch size equal to 13) into the memory\footnote{With careful optimisation of the training code and mixed precision multi-GPU training, the training time can be substantially reduced.}. We use AdamW optimiser~\cite{loshchilov2018decoupled} with base learning rate set to $1.3 \times 10^{-5}$. The learning rate is warmed up in the first 16K steps and then decays linearly. We set the weight decay to $10^{-4}$. During training, we let the model update the positional embeddings as \model{} needs to learn new positions not covered by \vlbert{} pre-training. The final model has $\sim$89.3M parameters excluding the word embeddings.

%%%%%%%%%%%%%%%%%%%%%
\paragraph{Decoding.}
%%%%%%%%%%%%%%%%%%%%%
At test time, we incrementally add the most likely prediction (\ie greedy search) into the previously masked position (Figure~\ref{fig:mmbert}) and shift the \T{[MASK]} token right by one. The reason we chose greedy over beam search is because the latter would make decoding much slower due to self-attentive representations being re-computed. The decoding ends when \T{[STOP]} is predicted.

%%%%%%%%%%%%%%%%%%%%%%%%%%%%%%
\subsection{Tasks \& Systems}
\label{sec:data_tasks}
%%%%%%%%%%%%%%%%%%%%%%%%%%%%%%
To evaluate \model{}'s generative abilities, we explore a diverse set of tasks: image captioning, text-only MT and multimodal MT. Table~\ref{tab:datasets} summarises the training statistics for the various datasets we use.

%%%%%%%%%%%%%%%%%%%%%%%%%%%%%%%%
\subsubsection{Image Captioning}
\label{sec:data_ic}
%%%%%%%%%%%%%%%%%%%%%%%%%%%%%%%%
Image captioning (IC) involves describing images in a specified natural language. We train \model{} for English, German and Turkish captioning tasks. Specifically, we use the \textsc{Flickr30k} dataset~\cite{young-etal-2014-image} that provides 29K training images, each with five \B{English} captions collected through crowd-sourcing. The validation and test sets
contain approximately 1K images each. We use the \textsc{Multi30k} dataset~\cite{elliott-etal-2016-multi30k}, which annotates \textsc{Flickr30k} images with five \B{German} captions. Finally, we use the \textsc{TasvirEt} dataset~\cite{unal2016tasviret} which provides two \B{Turkish} captions for each of the 8,092 images in the \textsc{Flickr8k} dataset~\cite{rashtchian-etal-2010-collecting}. Since \textsc{Flickr8k} is a subset of \textsc{Flickr30k}, we create a new split of \textsc{TasvirEt} to avoid data leakage between training and test splits. The resulting training, validation and test splits contain 6914, 543, and 543 images, respectively.

\begin{table}[t]
\centering
\resizebox{.9\columnwidth}{!}{%
\begin{tabular}{@{}lccrr@{}}
\toprule
\B{Name}  & \B{Type} & \B{Task}     & \B{Sents} & \B{Augm.} \\ \midrule
\textsc{Flickr8k}  & IC   & IM\ra TR   & 13,828    & 263K    \\
\midrule
\textsc{Multi30k}  & MMT  & DE\ra EN & 29,000      & 464K    \\
                   &      & FR\ra EN &             & 464K    \\
                   &      & EN\ra FR &             & 560K    \\
%\rowcolor{gray!10}
\textsc{Multi30k}  & MMT  & EN\ra DE & 29,000      & 582K    \\
\midrule
\textsc{Flickr30k} & IC   & IM\ra EN   & 145,000   & 2.39M  \\
                   &      & IM\ra DE   &           & 2.48M  \\
\midrule
\textsc{Iwslt}     & MT   & DE\ra EN & 158,388     & 3.85M  \\
                   &      & EN\ra DE &             & 4.43M  \\
\midrule
\textsc{Iwslt}     & MT   & FR\ra EN & 163,328     & 4.01M  \\
                   &      & EN\ra FR &             & 4.78M  \\
\midrule
\textsc{Setimes}   & MT   & TR\ra EN & 185,318     & 6.01M  \\
                   &      & EN\ra TR &             & 8.40M  \\
\bottomrule
\end{tabular}%
}
\caption{Training statistics of \model{}: the last column is the number of samples after sequence unrolling.}
\label{tab:datasets}
\end{table}
To evaluate \model{}'s performance on IC, we compare it against previous work with strong performance on COCO~\cite{chen2015microsoft} and \textsc{Flickr30k}. More precisely, \textsc{Adaptive Attention (Sentinel)}~\cite{adaptive_cap}, which uses a \I{sentinel} token to distinguish between visual and non-visual representations, and \textsc{Neural Baby Talk (Nbt)}, which follows a slot-filling approach through explicit object region information~\cite{lu2018neural}.

%%%%%%%%%%%%%%%%%%%%%%%%%%%%%%%%%%%%%%%%%%%%%%
\subsubsection{Multimodal Machine Translation}
\label{sec:data_mmt}
%%%%%%%%%%%%%%%%%%%%%%%%%%%%%%%%%%%%%%%%%%%%%%
Multimodal Machine Translation (MMT) attempts to improve MT quality by incorporating information from modalities other than language~\cite{sulubacak-etal-2020-multimodal}. In our case, we train \model{} for \textsc{En$\leftrightarrow$De} and \textsc{En$\leftrightarrow$Fr} MMT tasks and use the \textsc{Multi30k} dataset, the main dataset for image-informed translation, which provides caption translations for \textsc{Flickr30k} images in German and French. To evaluate \model{} on MMT tasks, we use the original \I{2016} test set which contains 1,000 examples.

For a comprehensive comparison with previous work, we train a SoTA recurrent MMT~ \cite{caglayan-etal-2020-simultaneous} \I{solely} on the \textsc{Multi30k} dataset, which applies a secondary (visual) attention in the decoder over the RoI features \ie the same features that are also used by \model{} ($\S$~\ref{sec:arch}). There are two GRU~\cite{cho-etal-2014-learning} layers in both the encoder and the decoder and the embedding \& hidden dimensions in the model are set to $200$ and $320$, respectively. Each model has $\sim$5.6M parameters excluding the word embeddings.

Besides the state-of-the-art \I{constrained} recurrent MMT model described above, we further compare \model{} -- which is trained on various other MT and IC corpora -- to an \I{unconstrained} Transformer-based MMT trained on $\sim$9M additional \textsc{En$\rightarrow$De} sentences~\cite{libovicky}\footnote{We obtained test set outputs from the author and pre-processed with \mbert{} tokeniser to ensure comparability.} in addition to \textsc{Multi30k}.

%%%%%%%%%%%%%%%%%%%%%%%%%%%%%%%%%%%%%%%%%%
\subsubsection{Text-only Machine Translation}
\label{sec:data_mt}
%%%%%%%%%%%%%%%%%%%%%%%%%%%%%%%%%%%%%%%%%%
We incorporate six text-only MT tasks into our training protocol. We use \textsc{En$\leftrightarrow$De} and \textsc{En$\leftrightarrow$Fr} MT datasets from IWSLT'14~\cite{cettolo-etal-2012-wit3} which consists of TED Talks' subtitles and their translations. We take the \T{prepare-iwslt14} recipe from \textsc{Fairseq}~\cite{ott-etal-2019-fairseq} to prepare the \I{dev} and \I{test} sets. This yields an \textsc{En$\leftrightarrow$De} test set of 6,750 sentences which consists of \I{dev2010, dev2012.TEDX, tst2010, tst2011 and tst2012}. Similarly, the \textsc{En$\leftrightarrow$Fr} test set consists of \I{dev2010, tst2010, tst2011 and tst2012}, which amounts to 4,493 sentences.

For \textsc{En$\leftrightarrow$Tr} directions, we use the SETIMES2~\cite{tiedemann-2012-parallel} news dataset for training.
For development and test sets, we take the official WMT test sets~\cite{bojar-etal-2018-findings}, namely, \I{newstest2016} and \I{newstest2017} as the development set (6,007 sentences), and \I{newstest2018} (6,000 sentences) as the test set.
Both IWSLT and SETIMES2 corpora are medium-scale resources often used in MT research community, and have much harder test sets than the MMT and IC tasks, due to a significant domain shift.

Finally, for each translation direction, we train a Transformer NMT model~\cite{vaswani2017attention} using the \textsc{Iwslt-De-En} recipe of the \textsc{Fairseq} toolkit~\cite{ott-etal-2019-fairseq}.
This recipe has six encoders and six decoders, each equipped with $4$-head self-attention layers. The model and feed-forward dimensions are set to $512$ and $1024$, respectively. Each model has $\sim$31.5M parameters excluding the word embeddings.
Since \model{} is a general purpose multilingual and multimodal generator, we expect it to perform in the same ballpark as these strong NMT baselines, but not necessarily be SoTA compared to novel \& sophisticated NMT models, which also make use of a lot more training data.
%%%%%%%%%%%%%%%%%%%%%%%%%%%%%%%

%%%%%%%%%%%%%%%%%%%%%%%%%%%%%%%%%
\section{Results and Findings} 
\label{sec:results}
We train \model{} on lowercased sentences for 45 epochs, after which the overall performance on the tasks reached a plateau. We define one \model{} epoch as a single pass over all of the training data for the \textsc{Multi30k En\ra De MMT} task and denote this task as the \I{reference task}. We use greedy search for all systems that we trained and merge back the \I{word pieces} before evaluation. We compute tokenised\footnote{Since \mbert{} is aggressive on splitting apostrophes and hyphens, our results may slightly differ from other work.}
BLEU~\cite{papineni-etal-2002-bleu}, METEOR~\cite{denkowski-lavie-2014-meteor} and CIDEr~\cite{vedantam2015cider} using \I{coco-caption}\footnote{https://github.com/tylin/coco-caption}.
In what follows, we provide detailed quantitative and qualitative findings.

%%%%%%%%%%%%%%%%%%%%%%%%%%%%%%%%
\subsection{Image Captioning}
\label{sec:res_img_cap}
%%%%%%%%%%%%%%%%%%%%%%%%%%%%%%%%
\begin{table}[t!]
\centering
\renewcommand{\arraystretch}{1.1}
\resizebox{.9\columnwidth}{!}{%
\begin{tabular}{rlrcc}
\toprule
\textsc{Task} &
\textsc{Approach} &
\textsc{Bl\,} &
\textsc{Mt} &
\textsc{Cr} \\
\midrule
\textsc{F30K En}   & \model{}             & 27.0       & \B{23.2}      & \B{0.587} \\
                   %& \citet{att_fcn}     & 23.0       & 18.9          & --        \\
                   & \textsc{Sentinel}\bs & 25.1       & 20.4          & 0.531     \\
                   & \textsc{Nbt}\bs      & \B{27.1}   & 21.7          & 0.575     \\
                   %& \citet{anderson2018bottom}& 27.3       & 21.7          & 0.566     \\
\midrule
\textsc{Coco En}   & \model{}       & 15.9       & 20.4          & 0.487     \\
                   & \textsc{Nbt}\bs          & \B{34.7}       & \B{27.1}          & \B{1.072}     \\
\hline
\rowcolor{gray!10}
\textsc{F30K Fr}   & \model{}       & 5.2        & 18.1          & 0.397     \\ 
\hline
\textsc{F8K Tr}    & \model{}             & 8.5         & 14.5          & 0.363     \\
\hline
\textsc{F30K De}   & \model{}             & 17.8        & 34.2          & 0.500     \\
%\hline

\bottomrule
\end{tabular}%
}
\caption{BLEU (\textsc{Bl}), METEOR (\textsc{Mt}) and CIDEr (\textsc{Cr}) scores for image captioning: gray background indicates \I{zero-shot} generation whereas \bs{} denotes the systems decoded with beam search.}
\label{tbl:ic_results}
\end{table}

Table~\ref{tbl:ic_results} provides an overview of \model{}'s image captioning performance on different test sets and languages. First of all, on \B{English} \textsc{Flickr30k}, \model{} is clearly able to outperform strong captioning models ($\S$~\ref{sec:data_ic}) \textsc{Sentinel}~\cite{adaptive_cap} and \textsc{Nbt}~\cite{lu2018neural}, even though they use beam search for decoding. On COCO~\cite{chen2015microsoft}, an image captioning corpus much larger and diverse than \textsc{Flickr30k}, we evaluate \model{} on Karpathy's test split~\cite{karpathy2015deep} and notice that the scores are reasonable given that \model{} \B{is not trained} on COCO: our model lags behind \textsc{Nbt} (w/ beam search) by 6.7 METEOR.

For \I{zero-shot} French captioning (\textsc{F30K Fr}), we resort to the reference MMT translations from the \textsc{Multi30k En\ra Fr} task, as there are no human references for French. Although this is problematic as the metrics will penalise captions that are not translations of English captions, we provide the scores to show that the \B{zero-shot} outputs are valid descriptions. We note that the low range of scores reported here is also due to having one reference caption instead of five references\footnote{As a reference, evaluating English captions using one reference at a time, yields 7.9 BLEU on average, compared to 27.0 BLEU in Table~\ref{tbl:ic_results}.} as in \textsc{Flickr30k}.
Finally we report results  for our custom \B{Turkish} split ($\S$~\ref{sec:data_ic}) (\textsc{F30K Tr}) and \B{German} (\textsc{F30K De}). Even though there are no comparable results in the literature for these three tasks, we demonstrate through some qualitative examples that \model{} produces sensible outputs.

\begin{table}[t!]
\renewcommand{\arraystretch}{1}
\centering
\resizebox{.99\columnwidth}{!}{%
\begin{tabular}{rl}
\toprule
\multicolumn{2}{c}{\includegraphics[height=2cm]{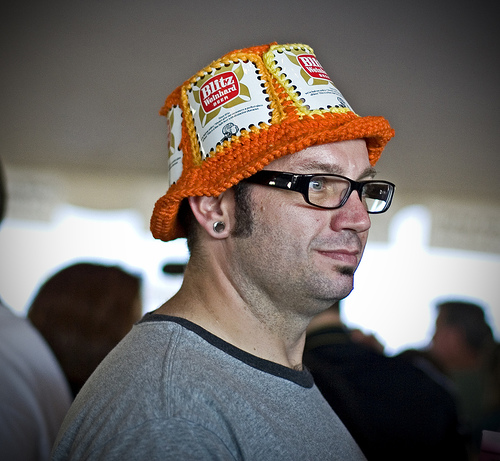}} \\
\textsc{En}     & a man wearing a hat and glasses.\\
\textsc{De}     & ein mann mit hut und brille. \\
                & \I{a man with hat and glasses}\\
\textsc{Tr}     & şapkalı ve gözlüklü bir adam.\\
                & \I{a man with a hat and glasses.}\\
\rowcolor{gray!10}
\textsc{Fr}     & un homme avec un chapeau et des lunettes. \\
\rowcolor{gray!10}
                & \I{a man with a hat and glasses.}\\
%\textsc{Mt Ref} & un homme avec un chapeau orange regardant quelque chose.\\
%\textsc{Ref En} & \I{a man with an orange hat staring at something.}
\hline
\\
\multicolumn{2}{c}{\includegraphics[height=2cm]{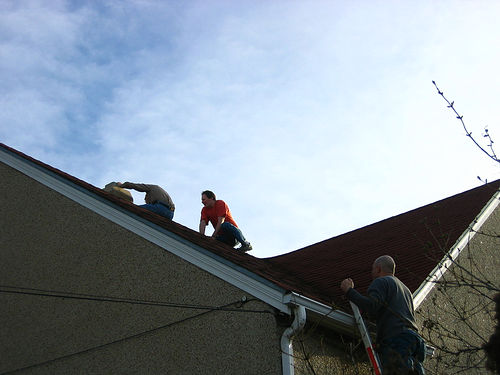}} \\
\textsc{En}     & two men are on a rooftop working on something. \\
\textsc{De}     & zwei männer arbeiten auf einem dach. \\
                & \I{two men working on a roof}\\
\textsc{Tr}     & iki binanın inşasında oturmuş, yanyana \\
                & yerde duran iki kişi.\\
                & \I{two people seated in the construction of two} \\
                & \I{buildings, standing next to each other on the ground. }\\
\rowcolor{gray!10}
\textsc{Fr}     & trois ouvriers du bâtiment construisent un toit. \\
\rowcolor{gray!10}
                & \I{three construction workers build a roof.}\\
\hline
\\
%\textsc{Mt Ref} & des gens réparent le toit d'une maison.\\
%\textsc{Ref En} & \I{people are fixing the roof of an house.}\\
\MC{2}{c}{\includegraphics[height=2cm]{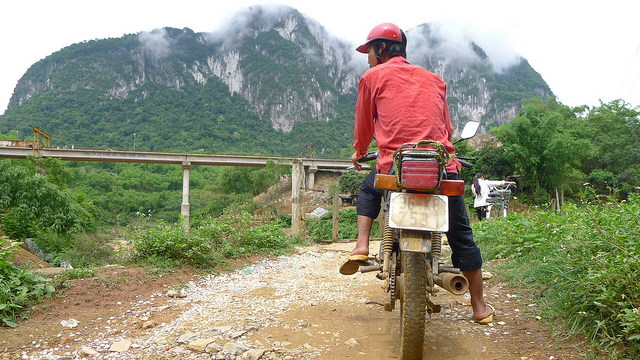}} \\
\textsc{En} & a man in a red shirt and helmet is riding a \\
            & motorbike on a dirt road. \\
\textsc{De} & ein mann fährt mit einem motorrad auf einem \\
            & weg an einem fluß entlang. \\
                & \I{a man rides a motorcycle on a path along a river.} \\
\textsc{Tr} & çamurlu bir yolda motoruyla ilerlemekte olan kırmızı \\
            & üstlü bir adam ve arkasındaki dağ manzarası. \\
                & \I{A man in a red top riding his bike down a muddy} \\
                & \I{road with a mountain landscape behind him.} \\
\rowcolor{gray!10}
\textsc{Fr} & un homme avec un casque fait du motocross. \\
\rowcolor{gray!10}           & 
     \I{a man with a helmet rides motocross.} \\
\bottomrule
    \end{tabular}}
\caption{Multilingual image captioning examples: The \I{italicised} sentences are Google Translate translations of \textsc{De,Tr,Fr} sentences into English. Gray background indicates \I{zero-shot} outputs. The last example is from \textsc{Coco} while the others are from \textsc{Flickr30k}.}
\label{tbl:cap_examples}
\end{table}

%%%%%%%%%%%%%%%%%%%%%%%%%%%%%%%%%
\paragraph{Qualitative examples.}
%%%%%%%%%%%%%%%%%%%%%%%%%%%%%%%%%
We now focus on a few examples to examine the multilingual image captioning ability of \model{} in action (Table~\ref{tbl:cap_examples}).
For the first image, all captions are almost the same as the image has few salient points. For the second image however, we observe much more variation across captions,  in line with the complexity of the scene.
We are particularly surprised by the \B{zero-shot} \B{French} captioning performance, a task that \model{} is not trained for at all. Upon manual inspection, we noticed that the captions are often short, objective gists of the images. These observations also hold for the captions generated for the COCO test set, as we can see in the third example. A set of additional examples in the Appendix shows that \model{} does not simply retrieve caption translations learned from the \textsc{En$\rightarrow$Fr} task.
Overall, both quantitative and qualitative results provide evidence of the utility of multimodal and multilingual initialisation as well as the efficacy of knowledge transfer across different tasks for image captioning.

\begin{table}[t!]
\centering
\resizebox{.9\columnwidth}{!}{%
\begin{tabular}{llcc}
\toprule
\textsc{mmt} &
\textsc{Approach} &
\textsc{Bl} &
\textsc{Mt} \\
\toprule
\textsc{en\ra de}  & \model{}\unc         & \B{42.2}            & \B{61.6}   \\
                   & \citet{libovicky}\unc\bs & 40.8                & 59.2       \\
                   & \citet{caglayan-etal-2020-simultaneous}    & 37.8       & 56.9       \\
                   & \textsc{Fairseq Nmt} & 37.5       & 56.1       \\
\midrule
\textsc{en\ra fr}  & \model{}\unc         & \B{68.0}   & \B{81.2}   \\
                   & \citet{libovicky}\unc\bs & 63.4       & 77.3       \\
                   & \textsc{Fairseq Nmt} & 61.5       & 75.5       \\
                   & \citet{caglayan-etal-2020-simultaneous}    & 61.0       & 75.3       \\
\hline
\rowcolor{gray!10}
\textsc{de\ra fr}  & \model{}\unc & \B{44.8}   & \B{64.1}   \\
                   & \citet{caglayan-etal-2020-simultaneous}     & 43.8       & 62.1       \\
                   & \textsc{Fairseq Nmt} & 41.7       & 60.7       \\
\hline
\rowcolor{gray!10}
\textsc{fr\ra de}  & \model{}\unc & \B{35.1}       & \B{56.9}   \\
                   & \textsc{Fairseq Nmt} & \B{35.1}       & 53.6       \\
                   & \citet{caglayan-etal-2020-simultaneous}     & 33.5       & 53.1       \\
\bottomrule
\end{tabular}%
}
\caption{BLEU (\textsc{Bl}) and METEOR (\textsc{Mt}) scores for MMT: \I{zero-shot} systems are highlighted with gray. Systems marked with \unc{} and \bs{} denote the use of auxiliary resources (\ie unconstrained) and beam-search decoding, respectively.} 
\label{tbl:mmt_results}
\end{table}

%%%%%%%%%%%%%%%%%%%%%%%%%%%%%%%%%%%%%%%%%%%
\subsection{Multimodal Machine Translation}
%%%%%%%%%%%%%%%%%%%%%%%%%%%%%%%%%%%%%%%%%%%
Table~\ref{tbl:mmt_results} summarises \model{}'s performance on MMT. First of all,
\model{} consistently outperforms the
Transformer-based \textsc{Fairseq} NMT models and the recurrent MMT~\cite{caglayan-etal-2020-simultaneous} models
on both the \textsc{En$\rightarrow$De} and the \textsc{En$\rightarrow$Fr} language pairs. Furthermore, \model{} is also substantially better than a state-of-the-art \I{unconstrained} MMT~\cite{libovicky} model trained on a  $\sim$6x larger parallel corpus. 

\paragraph{Adversarial evaluation.} Following 
\citet{elliott-2018-adversarial}, we probe \model{}'s ability for integrating multiple modalities effectively. Specifically, we decode translations by shuffling \{image, source caption\} mappings so that the images do not correspond to the sentences to be translated. The \textsc{En\ra De} results showed that the incongruence leads to $1.1$ and $0.9$ point \B{drops} in BLEU and METEOR, respectively. For \textsc{En\ra Fr}, the drops are much more prominent with $3.1$ and $2.3$ points again for BLEU and METEOR. This indicates that the features are not ignored at all, unlike in \cite{caglayan-etal-2019-probing}, where they showed that sequence-to-sequence MMT models can learn to ignore the images when the linguistic signal is sufficient to perform the task.

%%%%%%%%%%%%%%%%%%%%%%%%%%%%%%%%%%
\paragraph{Zero-shot performance.}
%%%%%%%%%%%%%%%%%%%%%%%%%%%%%%%%%%
The results in Table~\ref{tbl:mmt_results} show the surprising ability of \model{} to perform MMT on directions unseen during training. Moreover, the zero-shot performance surpasses strong MMT and NMT systems by up to 2 and 3.3 METEOR for \textsc{De\ra Fr} and \textsc{Fr\ra De}, respectively. Similar to the image captioning results, this demonstrates the potential of \model{} to generalise over a variety of language pairs and tasks.

%%%%%%%%%%%%%%%%%%%%%%%%%%%%%%%

\begin{table}[t]
\centering
\resizebox{.45\textwidth}{!}{%
\begin{tabular}{@{}lcccc@{}}
\toprule
& \MC{2}{c}{\textsc{Fairseq}} & \MC{2}{c}{\model{}} \\
\textsc{Task} &
\textsc{Bl} &
\textsc{Mt} &
\textsc{Bl} &
\textsc{Mt} \\
\midrule
\textsc{Iwslt En\ra De}   & 27.4     & 47.1    & \B{27.8} & \B{48.4} \\
\textsc{Iwslt De\ra En}   & 33.6     & 33.8    & \B{35.6} & \B{34.7} \\
\textsc{Iwslt En\ra Fr}   & \B{41.0} & 59.8    & 40.2     & \B{60.5} \\
\textsc{Iwslt Fr\ra En}   & 39.1     & 36.4    & \B{40.0} & \B{36.8} \\
\textsc{Setimes En\ra Tr} & \B{14.1} & 18.9    & 13.5     & \B{19.1} \\
\textsc{Setimes Tr\ra En} & 17.3     & 25.8    & \B{19.0} & \B{26.9} \\
\bottomrule
\end{tabular}%
}
\caption{Comparison of text-only MT performance of \model{} to each dedicated \textsc{Fairseq} NMT system: \model{} outperforms single models in most cases.}
\label{tab:mt_results}
\end{table}

%%%%%%%%%%%%%%%%%%%%%%%%%%%%%%%%
\subsection{Machine Translation}
%%%%%%%%%%%%%%%%%%%%%%%%%%%%%%%%
First, we compare \model{}'s performance to each task-specific \textsc{Fairseq} system. According to Table~\ref{tab:mt_results}, we observe that the translation quality of \model{} is generally superior compared to the strong \textsc{Fairseq} systems, especially in METEOR, where \model{} leads in all pairs.
%%%%%%%%%%%%%%%%%%%%%%%%
\begin{figure}[t!]
\centering
\includegraphics[width=.99\columnwidth]{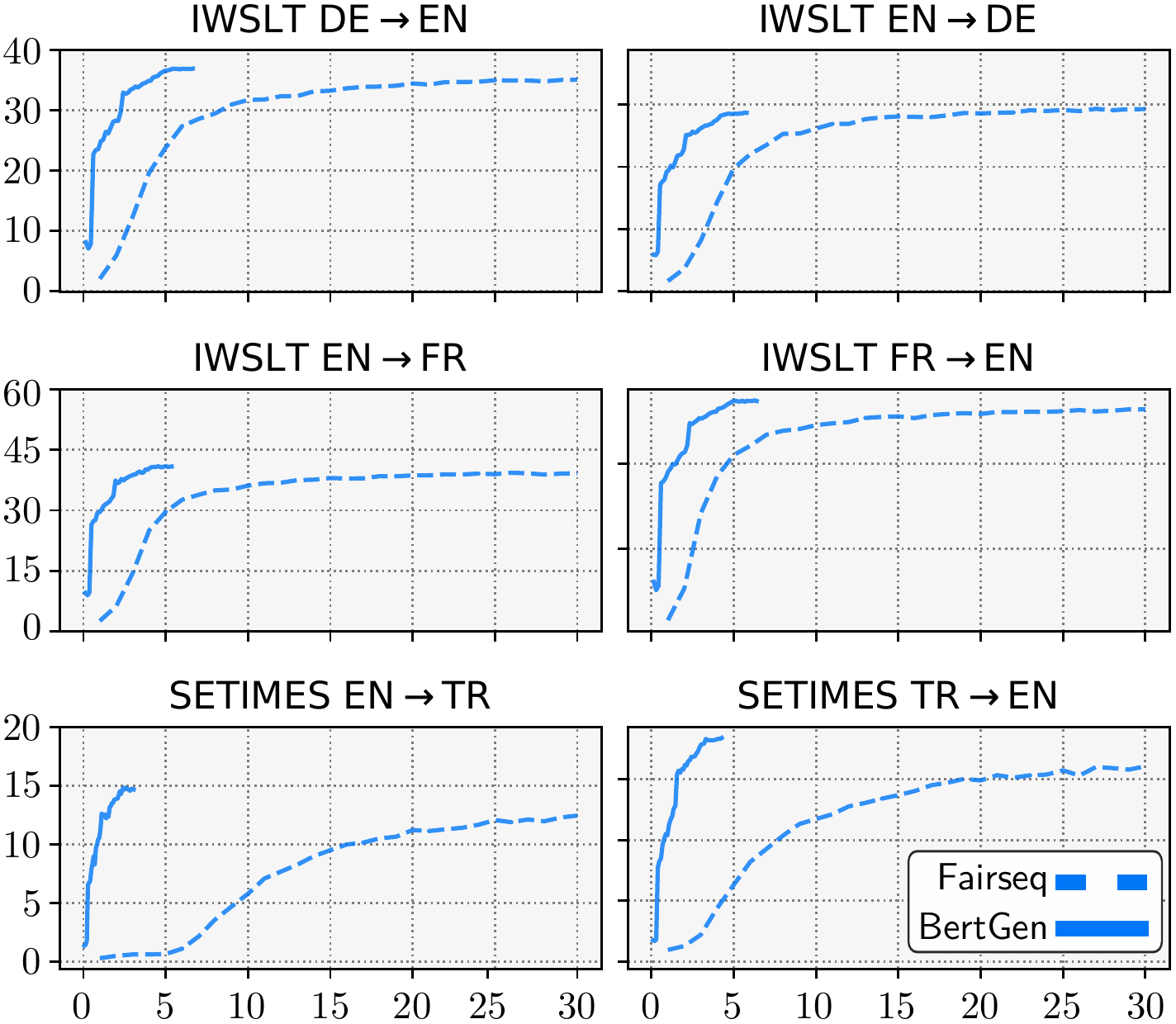}
\caption{\model{}'s learning efficiency on MT: validation scores are plotted against the number of full passes completed by \model{} and \B{each} \textsc{Fairseq} model, over the \I{corresponding} task's training set. Best checkpoints' test set performances are given in Table~\ref{tab:mt_results}.
}
\label{fig:val_plots}
\end{figure}
%%%%%%%%%%%%%%%%%%%%%%%%%

Second, we look at the learning efficiency by comparing the training curves between \model{} and \B{each} task-specific \textsc{Fairseq} system (Figure~\ref{fig:val_plots}).
Here, the x axis represents how many times the specific task's training set has been seen by the models. 
\model{} is trained for 45 reference epochs ($\S$~\ref{sec:results}),
and this corresponds to only a few complete passes over the training sets of NMT tasks\footnote{For example, only $\sim$3 passes over \textsc{Setimes En\ra Tr}.}.
This is in contrast to the single-task systems that usually require a large number of epochs for convergence. We notice a general trend and observe that \model{} tends to outperform single-task systems usually after only a few passes over the corresponding training set. Many factors could be contributing to this observation such as sequence unrolling, multi-tasking, shared input space or relevant inductive biases transferred from \mbert{}. 
We partly address these in the ablation studies ($\S$~\ref{sec:ablations}) and leave further investigation to future work.

%%%%%%%%%%%%%%%%%%%%%%%%%%%%%%%%%%
\paragraph{Zero-shot performance.}
%%%%%%%%%%%%%%%%%%%%%%%%%%%%%%%%%%
%%%%%%%%%%%%%%%%%%%%%%%%
\begin{table}[t]
\centering
\renewcommand{\arraystretch}{1.1}
\resizebox{.38\textwidth}{!}{%
\begin{tabular}{rcccc}
\toprule
 &
\MC{2}{c}{\textsc{De\ra Fr}} &
\MC{2}{c}{\textsc{Fr\ra De}} \\
       & \textsc{Bl} & \textsc{Mt}
       & \textsc{Bl} & \textsc{Mt}
\\
\hline
\rowcolor{gray!10}
\model{}             & 19.6     & 40.5       & 13.1   & 36.7  \\
\textsc{Tartu}\bs    & 39.5     & 59.0       & 26.3   & 47.3  \\
\textsc{Msra}\bs     & 46.5     & 64.2       & 38.2   & 56.4  \\
\bottomrule
\end{tabular}%
}
\caption{\I{Zero-shot} \model{} performance on WMT'19 test set: TARTU and MSRA systems are not \I{zero-shot} as they are trained on \textsc{De$\leftrightarrow$FR} corpora. Systems marked with \bs{} are beam search outputs.}
\label{tab:zeroshot_wmt}
\end{table}

We use the \textsc{De$\leftrightarrow$Fr} test set from the WMT'19 \I{shared task on news translation}~\cite{barrault-etal-2019-findings} to assess the \B{zero-shot translation} capability of \model{}. This test set includes 1,701 sentences from news data regarding \I{European Elections}. We compare our results to two shared task systems, namely \textsc{Tartu} (baseline) and \textsc{Msra} (state-of-the-art)~\cite{barrault-etal-2019-findings}, after re-tokenising them accordingly with \mbert{}\footnote{\textsc{Tartu} is the baseline and \textsc{Msra} is the best performing system for the shared task}.
%%%%%%%%%%%%%%%%%%%%%%%%
\begin{table*}[t!]
\centering
\resizebox{.98\textwidth}{!}{%
\begin{tabular}{rl}
\toprule
\model{}:        & la décision est tombée au 70ème anniversaire de ma femme. \\
& \I{the decision fell on my wife's 70th birthday.} \\
\textsc{Wmt Ref} & la décision est tombée le jour du 70ème anniversaire de ma femme. \\
& \I{the decision fell on my wife's 70th birthday.} \\
\midrule
\model{}:        & en espagne, on s'est malheureusement habitué à une rôle double et passive. \\
& \I{in spain, we unfortunately got used to a \B{double} and passive role.} \\
\textsc{Wmt Ref} & en espagne, on s'est malheureusement habitué à un rôle secondaire, passif. \\
& \I{in spain, we unfortunately got used to a \B{secondary}, passive role.} \\
\midrule
\model{}:        & pas parce que le président du fdp a dit quelque chose qu' ils ont défaillant leur vote. \\
& \I{not because the fdp president said something that they \B{missed their vote.}} \\ 
\textsc{Wmt Ref} & ce n' est pas parce que le président fédéral du fdp a dit quelque chose qu' ils ont refusé d' approuver. \\
& \I{it is not because the federal president of the fdp said something that they \B{refused to approve.}}\\
\bottomrule
\end{tabular}}%
\caption{Zero-shot \textsc{De$\rightarrow$Fr} translations on WMT'19 test set. The \I{italicised} sentences are Google Translate translations of French sentences into English. \B{Bold} indicates important differences between \model{} and references.}
\label{tab:defr_examples_app}
\end{table*}
%%%%%%%%%%%%%%%%%%%%%%%%
Although \model{} is expected to obtain lower scores than the dedicated WMT systems due to the domain mismatch of the test set, we consider both the quantitative (Table~\ref{tab:zeroshot_wmt}) and the qualitative results (Table~\ref{tab:defr_examples_app}) extremely encouraging.

%%%%%%%%%%%%%%%%%%%%%%%%%%%
\subsection{Ablation Studies}
\label{sec:ablations}
%%%%%%%%%%%%%%%%%%%%%%%%%%%
\subsubsection{Impact of initialisation}
\label{sec:ablat_init}
\begin{figure}[t!]
\centering
\includegraphics[width=.9\columnwidth]{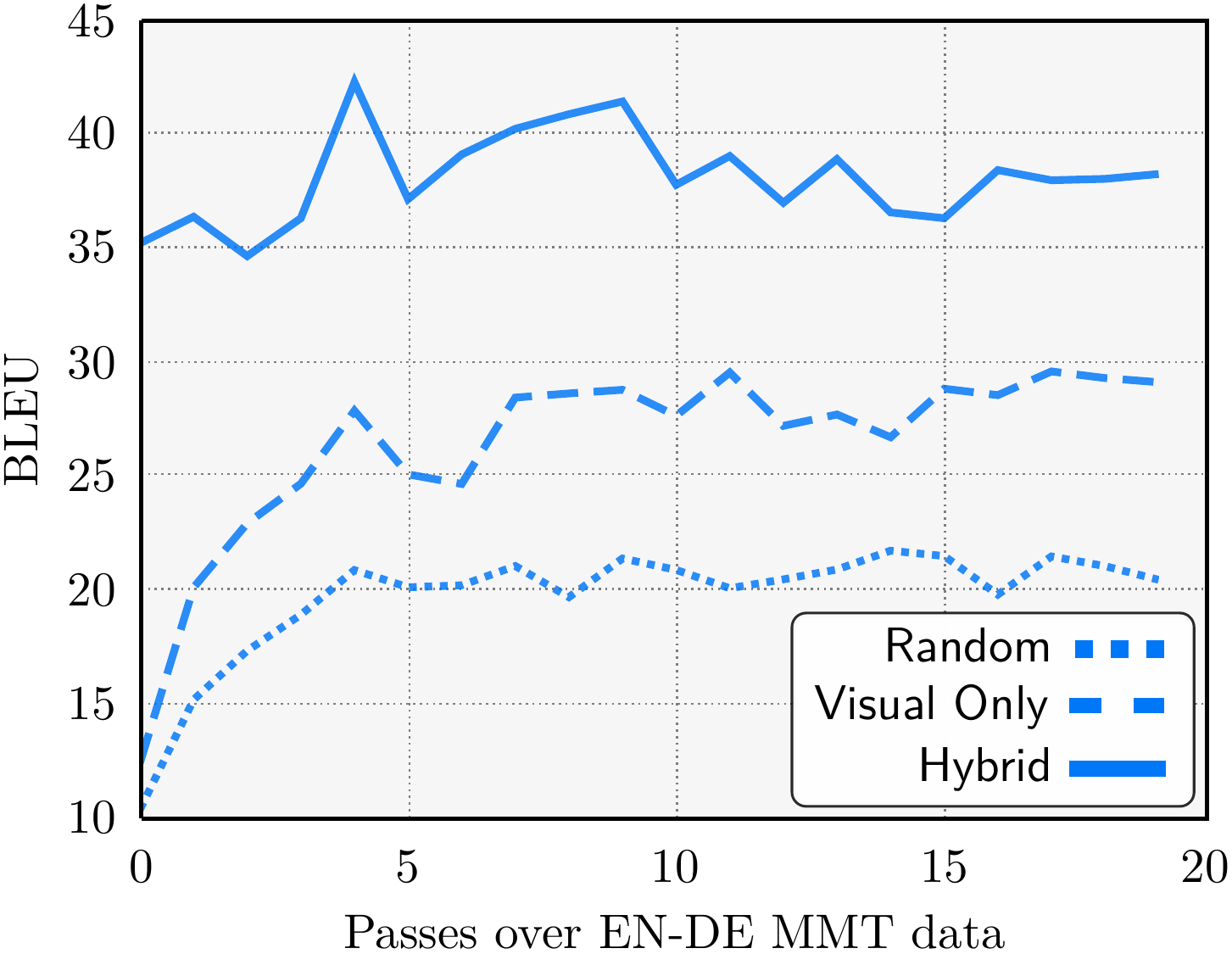}
\caption{Validation scores on \textsc{Multi30k} \textsc{En$\rightarrow$De} MMT for the \B{initialisation} ablation: Hybrid initialisation is the most beneficial strategy for \model{}.}
\label{fig:ablat_init}
\end{figure}

We train \I{single-task} MMT systems on the \textsc{Multi30k} \textsc{En$\rightarrow$De} language pair. Specifically, we begin with a baseline system which is initialised with \B{random} weights. We then train a second baseline where only the \B{visual} processing layers are transferred from \vlbert{}. Finally, we train a third baseline that is initialised similar to \model{}, \ie using the \B{hybrid} initialisation ($\S$~\ref{sec:method_init}). Figure~\ref{fig:ablat_init} compares the validation BLEU scores of these three systems. We observe that the benefits of knowledge transfer from pre-trained models are incrementally positive, however, \model{}'s hybrid initialisation outperforms the other two ablations.

%%%%%%%%%%%%%%%%%%%%%%%%%%%%%%%%%%%%%%%%%%%%%
\subsubsection{Impact of multi-task training}
\begin{figure}[t]
\centering
\includegraphics[width=.9\columnwidth]{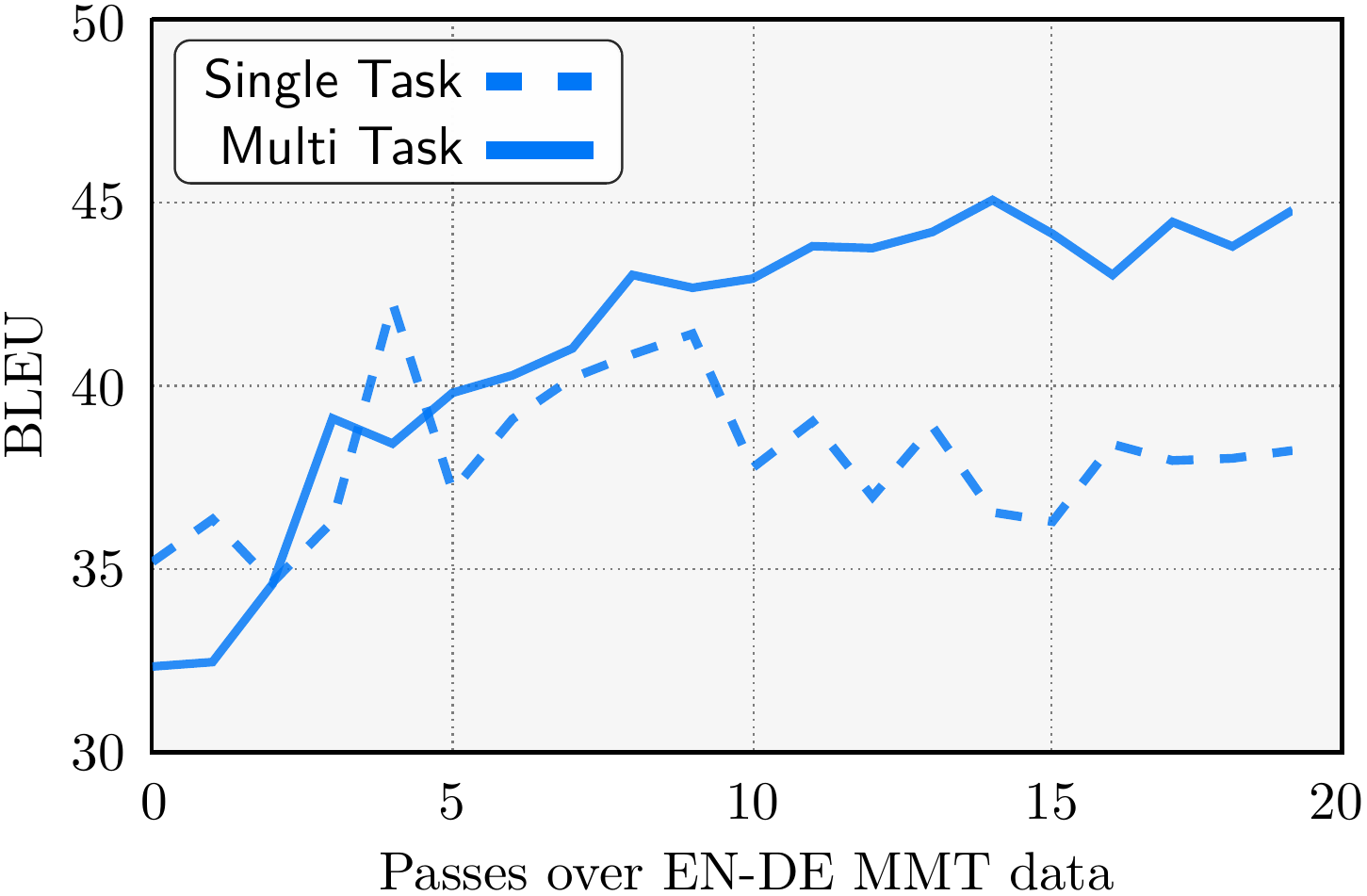}
\caption{Validation scores on \textsc{Multi30k} \textsc{En$\rightarrow$De} MMT for the \B{multi-tasking} ablation: The default multi-task \model{} outperforms the single-task one.}
\label{fig:ablat_mtl}
\end{figure}

%%%%%%%%%%%%%%%%%%%%%%%%%%%%%%%%%%%%%%%%%%%%%
\label{sec:ablat_mtl}
We now remove the multi-tasking aspect from \model{} to investigate the extent to which the performance improvements are related to other tasks. Similar to $\S$~\ref{sec:ablat_init}, we focus on the \textsc{Multi30k} \textsc{En$\rightarrow$De} MMT task and train a single-task, \I{hybrid-initialised} \model{}. Figure~\ref{fig:ablat_mtl} compares the validation BLEU scores obtained by the default \model{} and the single-task variant. We observe that \model{} benefits from multi-task training and, more importantly, does not seem to exhibit patterns of catastrophic forgetting~\cite{french-1999-catastrophic}. Based on these observations, we expect similar model behavior to hold for other tasks.

%%%%%%%%%%%%%%%%%%%%%%%%%%%%%%%%%

%%%%%%%%%%%%%%%%%%%%%%
\section{Related Work}
\label{sec:related}

\subsection{Multimodal multilingual pre-training}
%%%%%%%%%%%%%%%%%%%%%%%%%%%%%%%%%%%%%%%%%%%%%%%%%%%
Research in NLP and related fields has been increasingly focusing on transfer learning  approaches where a model is first pre-trained on a data-rich task, and then transferred to downstream tasks~\cite{mccann2017learned,peters-etal-2018-deep,devlin-etal-2019-bert}. This framework presumably allows the model to capture useful inductive biases that generalise to a variety of NLP tasks, often after performing a task-specific fine-tuning~\cite{raffel2019exploring}.
Of these, the most relevant studies to our work are BERT~\cite{devlin-etal-2019-bert} and its multilingual version \mbert{}, which pre-train a Transformer~\cite{vaswani2017attention}  on large monolingual corpora using the masked language modelling (MLM) objective.

Recent research has also attempted to combine linguistic inputs with other modalities such as vision and speech, to achieve a grounded understanding of meaning. Successful approaches including LXMERT~\cite{tan-bansal-2019-lxmert}, VL-BERT~\cite{su2019vl} and others~\cite{lu2019vilbert,unicodervl,oscar} achieve this by combining BERT's MLM objective with auxiliary tasks such as masked region classification and image sentence matching, and pre-train their model on large-scale image captioning corpora~\cite{chen2015microsoft,sharma-etal-2018-conceptual}. Similarly, SpeechBERT extends BERT by jointly training on speech and text data~\cite{chuang2019speechbert}.
Although SoTA results are reported by these approaches, they focus on unimodal and multimodal natural language understanding (NLU) tasks, with a strong emphasis in English. The backbone of \model{} combines VL-BERT~\cite{su2019vl} with \mbert~\cite{devlin-etal-2019-bert} to realise \I{a multilingual and multimodal generator} that can be used for a diverse set of generative tasks and languages rather than NLU tasks.

%%%%%%%%%%%%%%%%%%%%%%%%%%%%%%%%%%%%%%%%%%%%%%%%
\subsection{Pre-training for generative tasks}
%%%%%%%%%%%%%%%%%%%%%%%%%%%%%%%%%%%%%%%%%%%%%%%%
Previous work has studied how to benefit from pre-trained BERT models in generative tasks such as NMT~\cite{imamura-sumita-2019-recycling,clinchant-etal-2019-use,incorporating_bert}. \model{} differs from these as it is not fine-tuned for a particular MT corpus and it exhibits multi-lingual and multi-modal properties for general purpose generation.

Another related branch of work explores pre-training strategies specific to sequence-to-sequence tasks. This includes  MASS~\cite{song2019mass}, which exploits an encoder-decoder framework with the MLM objective for task-specific generative pre-training and UniLM~\cite{dong-etal-2019-unilm}, which introduces uni-directional, bi-directional and sequence-to-sequence LM objectives by carefully adjusting the self-attention masks during training. \citet{Zhou_Palangi_Zhang_Hu_Corso_Gao_2020} extend UniLM to vision \& language pre-training using Conceptual Captions~\cite{sharma-etal-2018-conceptual} as the pre-training dataset. However, these models require a further fine-tuning step for generative tasks, unlike \model{} that is trained only \I{once}.

%%%%%%%%%%%%%%%%%%%%%%%%%%%%%%%%%%%%%%%%%%%%%%
\subsection{Multi-task learning for generation}
%%%%%%%%%%%%%%%%%%%%%%%%%%%%%%%%%%%%%%%%%%%%%%
Several approaches exist for multi-task learning \& generation~\cite{dong-etal-2015-multi,luong-etal-2016-multi} in NLP, especially in multilingual NMT, where tasks denote different language pairs~\cite{zoph-knight-2016-multi,firat-etal-2016-multi}.
The multi-task (and zero-shot) generation ability of \model{} is mostly inspired by
\citet{ha-etal-2016-toward} and \citet{johnson-etal-2017-googles}. Both of these introduced target language specifiers to select the output language when decoding translations from their model.

Our multilingual \& multimodal take on multi-task generation is most similar to \citet{kaiser-etal-2017-one}, where a single Transformer model is trained on different tasks including image captioning, object classification, machine translation, speech recognition and parsing. However, their architecture depends on particular structures such as encoders, decoders, modality-specific networks and I/O mixers, unlike \model{} which does not require task-specific modules.

%%%%%%%%%%%%%%%%%%%%%
\section{Conclusions}
\label{sec:conclusion}
In this paper, we presented \model{}, a novel generative, decoder-only model which extends BERT by combining multimodal and multilingual pre-trained models. Our findings show that \model{} obtains strong performance on a variety of generative tasks and further generalises over unseen tasks. Importantly, our model demonstrates the potential for general-purpose (instead of task-specific) generation that is above and beyond the traditional pre-training and fine-tuning practices. \model{} is also parameter efficient as it has 89.3M total parameters and is trained on thirteen tasks encompassing MT, multimodal MT and image captioning. On the other hand, each of the single-task \textsc{Fairseq} NMT baselines has 31.5M parameters.

Our ablation studies show that \model{} is able to efficiently transfer relevant inductive biases from the pre-trained models and benefits from multi-task learning without suffering from catastrophic forgetting. We hope that these findings will motivate future research in exploiting more sophisticated pre-trained models in place of \mbert{} and \vlbert{} and others.

%%%%%%%%%%%%%%%%%%%%%

\section*{Acknowledgments}
This paper is a follow-up work to the MSc. Thesis of Faidon Mitzalis, co-supervised by Prof. Lucia Specia and Dr. Ozan Caglayan.
Lucia Specia, Pranava Madhyastha and Ozan Caglayan received support from MultiMT project (H2020 ERC Starting Grant No. 678017). Lucia Specia also received support from the Air Force Office of Scientific Research (under award number FA8655-20-1-7006).

\bibliographystyle{acl_natbib}
\bibliography{anthology,acl2021}

\onecolumn
\appendix
\section{Qualitative Examples}

% \begin{table}[t]
% \centering
% \renewcommand{\arraystretch}{1.0}
% \resizebox{.45\textwidth}{!}{%
% \begin{tabular}{lcccc}
% \toprule
%  &
% \MC{2}{c}{\textsc{en\ra de}} &
% \MC{2}{c}{\textsc{en\ra fr}} \\
%       & \textsc{bl} & \textsc{mt}
%       & \textsc{bl} & \textsc{mt}
% \\ \midrule
% \model{}                & 42.2     & 61.6       & 68.0     & 81.2  \\
% %                       & 41.1     & 60.7       & 64.9     & 78.9
% \,\,+\textsc{shuffle}   & \da{1.1} & \da{0.9}   & \da{3.1} & \da{2.3}\\
% \bottomrule
% \end{tabular}%
% }
% \caption{Incongruent MMT decoding: when image order is shuffled across samples in the test set, \model{} scores drop substantially.}
% \label{tbl:mmt_incongruence}
% \end{table}

%%%%%%%%%%%%%%%%%%%%
% FLICKR FR EXAMPLES
%%%%%%%%%%%%%%%%%%%%
\vspace*{9em}
\begin{table*}[h]
\renewcommand{\arraystretch}{1}
\centering
\resizebox{.98\textwidth}{!}{%
\begin{tabular}{cl}
\toprule
\multirow{4}{*}[-1.2em]{\includegraphics[width=2.8cm]{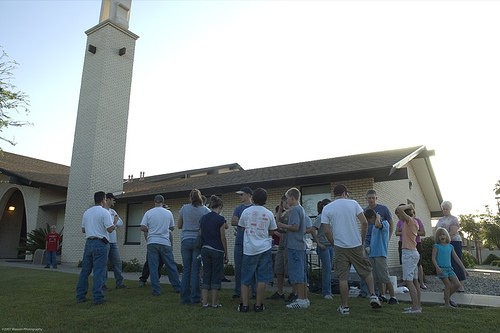}} &
\\
& \model{}: un groupe de jeunes sont réunis, et ils sont debout à l 'extérieur d'un bâtiment. \\
& \phantom{\model{}:} (a group of young people are gathered, and they are standing outside a building.) \\
& \textsc{\phantom{w}Mt Ref:} des gens debout devant un bâtiment.\\
& \phantom{w\textsc{Mt Ref:}} (people standing outside of a building.)\\
\\ \midrule
% TR bir grup insan binanın önünde duruyor.
% DE eine gruppe von menschen steht vor einem gebäude.
% EN a group of people standing in front of a building .

\multirow{4}{*}[-1.2em]{\includegraphics[width=2.8cm]{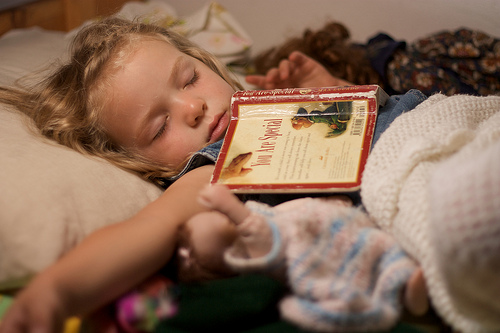}} &
\\
& \model{}: une petite fille lit un livre.\\
& \phantom{\model{}:} (a little girl reads a book.) \\
& \textsc{\phantom{w}Mt Ref:} un jeune enfant dormant dans son lit avec un livre ouvert sur sa poitrine.\\
& \phantom{w\textsc{Mt Ref:}} (a young child sleeping in her bed with an open book on her chest.) \\
\\ \midrule
% TR bir çocuk okuyor .
    %  -> a child is reading. 
% DE ein kleines mädchen liest in einem buch .
    % -> a little girl is reading a book.  
% EN a young girl is reading a book while laying on a blanket . .

\multirow{4}{*}[-1.2em]{\includegraphics[width=2.8cm]{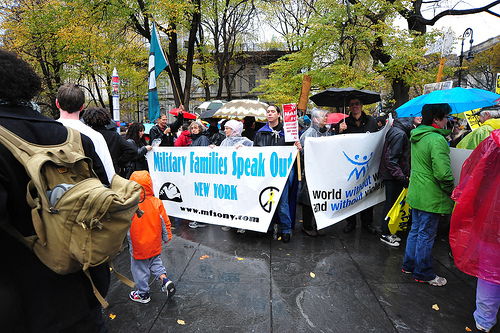}} &
\\
& \model{}: des manifestants avec des pancartes. \\
& \phantom{\model{}:} (demonstrators with placards.) \\
& \textsc{\phantom{w}Mt Ref:} des familles de militaires défilent dans new york un jour de pluie. \\
& \phantom{w\textsc{Mt Ref:}} (military families are marching through new york on a rainy day.) \\
% TR elinde bir pankartla sokakta eylem yapan bir adam .
    % -> a man doing action in the street with a banner in his hand. 
% DE eine gruppe von menschen steht mit plakaten auf einer straße.
    % -> a group of people is standing on a street with posters.  
% EN a man in a brown jacket is holding a sign that says " free hugs " .
 
\\
\bottomrule
\end{tabular}}
\caption{Zero-shot French image captioning examples for the \textsc{Flickr30k} test set.}
\label{tbl:fr_examples_app}
\end{table*}

%%%%%%%%%%%%%%%
% COCO EXAMPLES
%%%%%%%%%%%%%%%

\begin{table*}[t!]
\centering
\resizebox{.85\textwidth}{!}{%
\begin{tabular}{@{}rl@{}}
\toprule
\MC{2}{c}{\includegraphics[width=3.5cm]{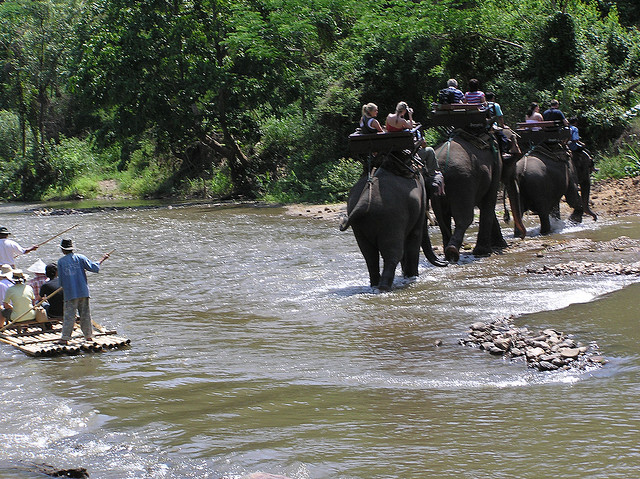}} \\
EN: & a group of people are riding on \B{elephants} through a river. \\
FR: & un groupe de personnes sur des chevaux sur un bateau dans un ruisseau. \\
    & \I{a group of people on \B{horses} on a boat in a stream.} \\
DE: & eine gruppe reiter fährt auf einem fluss. \\
    & \I{a group of riders is riding on a river.} \\
TR: & bir grup insan bir derede duran dört tane at ile ilerliyorlar. \\
    & \I{a group of people are moving with four \B{horses} standing in a stream.} \\
\midrule
\MC{2}{c}{\includegraphics[width=3.5cm]{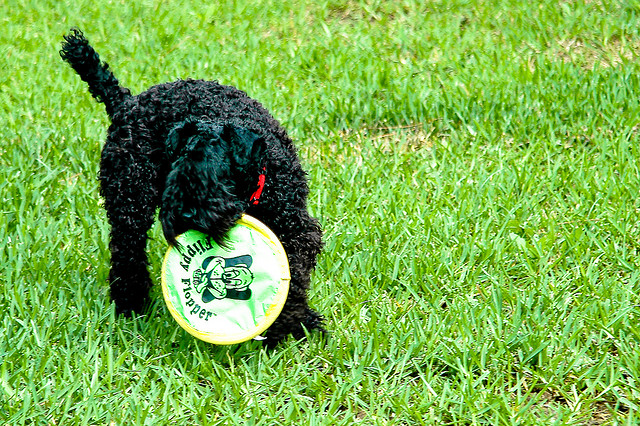}} \\
EN: & a \B{black} dog is playing with a \B{yellow toy} in the grass. \\
FR: & un chien avec une frisbee dans la pelouse. \\
    & \I{a dog with a \B{frisbee} in the lawn.} \\
DE: & ein schwarzer hund mit rotem halsband spielt mit einem gelben ball auf einer wiese. \\
    & \I{a \B{black} dog with a \B{red collar} is playing with a \B{yellow ball} in a meadow.} \\
TR: & yeşil bir topu ısırmaya çalışan siyah bir köpek.\\
    & \I{a \B{black} dog trying to bite a \B{green ball.}} \\
\midrule
\MC{2}{c}{\includegraphics[width=3.5cm]{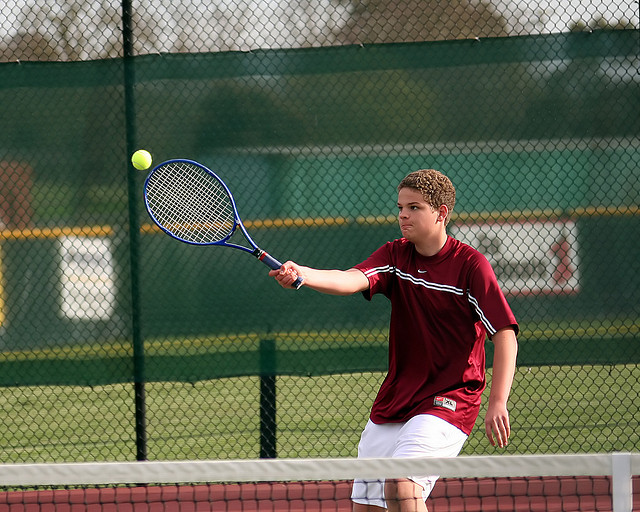}} \\
EN: & a boy in a red shirt and white shorts is playing tennis. \\
FR: & un \B{tennisteur} frappe une balle. \\
    & \I{a \B{tennisteur} hits a ball.} \\
DE: & ein junge spielt tennis. \\
    & \I{a boy is playing tennis.}\\
TR: & tenis raketi ile topa vuran çocuk. \\
    & \I{boy hitting the ball with a tennis racket.} \\
\bottomrule
\end{tabular}}
\caption{COCO captioning examples: \I{Italicised captions are Google's translations into English for non-English examples}. \B{Bold} phrases highlight lexical variations or errors related to the salient visual concepts in the images. The last example shows a morphological error that \model{} does when trying to generate a \B{tennis player} in French.} 
\label{tbl:coco_examples}
\end{table*}

\begin{table*}[t!]
\centering
\resizebox{.8\textwidth}{!}{%
\begin{tabular}{@{}rl@{}}
\toprule
\model{}         & mehrere ngos, darunter die mozilla - und greenpeace - stiftung,\\ 
                 & schätzen, dass diese neuen werkzeuge unfähig sind und zu spät kommen. \\ \cmidrule(l){2-2}
                 & \I{several ngos, including the mozilla and greenpeace foundations, } \\
                 & \I{estimate that these new tools are \B{incapable} and come too late.} \\ \cmidrule(l){2-2}

\textsc{Wmt Ref} & mehrere ngos, unter denen die mozilla - stiftung und greenpeace, \\
                 & schätzen, dass diese neuen tools unzureichend sind und zu spät kommen. \\ \cmidrule(l){2-2}
                 & \I{several ngos, including the mozilla foundation and greenpeace,} \\
                 & \I{estimate that these new tools are \B{inadequate} and come too late.} \\
\midrule

\model{}         & immigration wird als ein großes problem für die ue betrachtet,\\
                 & für 45 prozent der deutschen und 40 prozent aller europäischen. \\ \cmidrule(l){2-2}
                 & \I{immigration is seen as a big problem for the \B{ue}, } \\
                 & \I{for 45 percent of germans and 40 percent of all european ones.} \\ \cmidrule(l){2-2}
\textsc{Wmt Ref} & die einwanderung halten 45 prozent der deutschen und 40 prozent \\
                 & aller europäer für das größte problem der eu. \\ \cmidrule(l){2-2}
                 & \I{45 percent of germans and 40 percent of all europeans} \\                    & \I{consider immigration to be the biggest problem in the \B{eu}.}\\
\midrule

\model{}         & das ist der grund, warum er in seinem buch die frage erforscht, \\
                 & ob es alternativen zu wahlen gibt. \\ \cmidrule(l){2-2}
                 & \I{that is the reason why he explores the question of whether} \\
                 & \I{there are alternatives to choose from in his book.} \\ \cmidrule(l){2-2}
\textsc{Wmt Ref} & deshalb geht er in seinem buch der frage nach, \\
                 & ob es alternativen zu wahlen gibt. \\ \cmidrule(l){2-2}
                 & \I{therefore, in his book, he investigates the question of whether} \\
                 & \I{there are alternatives to choose from.}\\
\bottomrule
\end{tabular}}
\caption{Zero-shot \textsc{Fr$\rightarrow$De} translations of WMT'19 test set. The \I{italicised} sentences are Google Translate translations of the German outputs into English.}\label{tbl:frde_examples_app}
\end{table*}

\end{document}